%% file: paper.tex
\documentclass[final]{cvpr}

\usepackage{times}
\usepackage{epsfig}
\usepackage{graphicx}
\usepackage{amsmath}
\usepackage{amssymb}

\usepackage{xcolor}
\usepackage{booktabs}
\usepackage{subfigure}
\makeatletter
\@namedef{ver@everyshi.sty}{}
\makeatother
\usepackage{pgf}
\usepackage{xspace}
\usepackage{soul}
\usepackage{tikz}
\usetikzlibrary{positioning}
\usetikzlibrary{matrix}
\usetikzlibrary{arrows}
\usetikzlibrary{calc}
\usetikzlibrary{3d,decorations.text,shapes.arrows,fit,backgrounds}
\usetikzlibrary{decorations.pathreplacing,calligraphy}
\usetikzlibrary{angles,quotes}

\usepackage[pagebackref=true,breaklinks=true,colorlinks,bookmarks=false]{hyperref}
\usepackage{microtype}
\usepackage{cleveref}

\def\cvprPaperID{7161} %
\def\confYear{CVPR 2021}

\begin{document}

\graphicspath{{figures/}}
\DeclareGraphicsExtensions{.pdf,.jpeg,.png}

\newcommand{\methodname}{LaneATT}
\newcommand\sbullet[1][.5]{\mathbin{\vcenter{\hbox{\scalebox{#1}{$\bullet$}}}}}

\title{Keep your Eyes on the Lane: Real-time Attention-guided Lane Detection}

\author{
	Lucas Tabelini$^1$ ,  Rodrigo Berriel$^1$, Thiago M. Paixão$^2$,\\
	Claudine Badue$^1$, Alberto F. De Souza$^1$, Thiago Oliveira-Santos$^1$ \vspace{0.25ex} \\
	$^1$Universidade Federal do Espírito Santo (UFES) ~~$^2$Instituto Federal do Espírito Santo (IFES)\\
    \footnotesize{\texttt{tabelini@lcad.inf.ufes.br}} \\
}

\maketitle

\input{secs/0_abstract.tex}
\input{secs/1_introduction.tex}

\input{secs/2_relatedwork.tex}
\input{secs/3_method.tex}

\input{secs/4_experiments.tex}

\input{secs/5_conclusion.tex}

{\small
\bibliographystyle{ieee_fullname}
\bibliography{refs}
}

\end{document}

%% file: secs/0_abstract.tex
\begin{abstract}
Modern lane detection methods have achieved remarkable performances in complex real-world scenarios, but many have issues maintaining real-time efficiency, which is important for autonomous vehicles. In this work, we propose \methodname: an anchor-based deep lane detection model, which, akin to other generic deep object detectors, uses the anchors for the feature pooling step. Since lanes follow a regular pattern and are highly correlated, we hypothesize that in some cases global information may be crucial to infer their positions, especially in conditions such as occlusion, missing lane markers, and others. Thus, this work proposes a novel anchor-based attention mechanism that aggregates global information. The model was evaluated extensively on three of the most widely used datasets in the literature. The results show that our method outperforms the current state-of-the-art methods showing both higher efficacy and efficiency. Moreover, an ablation study is performed along with a discussion on efficiency trade-off options that are useful in practice.

\end{abstract}

\vspace{-12pt}

%% file: secs/1_introduction.tex
\section{Introduction}

Deep learning has been essential for recent advances in numerous areas, especially in autonomous driving~\cite{badue2020self}. Many of the deep learning applications in self-driving cars are in their perception systems. To be safe around humans, autonomous vehicles should perceive their surroundings, including the position of other vehicles and themselves. In the end, the more predictable a car's movement is, the safer it will be for its passengers and pedestrians. Thus, it is important for autonomous vehicles to know each lane's exact position, which is the goal of lane detection systems.

Lane detection models have to overcome various challenges. A model that will be used in a real-world scenario should be robust to several adverse conditions, such as extreme light and weather conditions. Moreover, lane markings can be occluded by other objects (\eg, cars), which is extremely common for self-driving cars.
Some approaches, such as polynomial regression models, may also suffer from a data imbalance problem caused by the long-tail effect since cases with sharper curves are less common.
Besides, the model not only has to be robust but also efficient. In several applications, lane detection must perform in real-time, or faster to save processing power for other systems, a requirement that many models struggle to cope with.

There are numerous works in the literature that tackle this problem. Before the advent of deep learning, several methods used more traditional computer vision techniques, such as Hough lines~\cite{berriel2017ego,assidiq2008real}. More recently, focus has shifted to deep learning approaches with the advance of convolutional neural networks (CNNs)~\cite{scnn,enet-sad,fastdraw}. In this context, the lane detection problem is usually formulated as a segmentation task, where, given an input image, the output is a segmentation map with per-pixel predictions~\cite{scnn}. Although recent advances in deep learning have enabled the use of segmentation networks in real-time~\cite{romera2017erfnet}, various models struggle to to achieve real-time performance. Consequently, the number of backbone options for segmentation-based methods is rather limited. Hence, some recent works have proposed solutions in other directions~\cite{linecnn,polylanenet}. Apart from that, many other issues are common in works on lane detection, such as the need for a post-processing step (usually a heuristic), long training times, and a lack of publicly available source code, which hinders comparisons and reproducibility.

In this work, we present a method for real-time lane detection that is both faster and more accurate than most state-of-the-art methods. We propose an anchor-based single-stage lane detection model called \methodname. Its architecture enables the use of a lightweight backbone CNN %
while maintaining high accuracy. A novel anchor-based attention mechanism to aggregate global information is also proposed. Extensive experimental results are shown on three benchmarks (TuSimple~\cite{tusimple}, CULane~\cite{scnn} and LLAMAS~\cite{llamas}), along with a comparison with state-of-the-art methods, a discussion on efficiency trade-offs, and an ablation study of our design choices. In summary, our main contributions are:

\begin{itemize}
    \item A lane detection method that is more accurate than existing state-of-the-art real-time methods on a large and complex dataset;
    \item A model that enables faster training and inference times than most other models (reaching 250 FPS and almost an order of magnitude less multiply-accumulate operations (MACs) than the previous state-of-the-art);
    \item A novel anchor-based attention mechanism for lane detection which is potentially useful in other domains where the objects being detected are correlated.
\end{itemize}

%% file: secs/2_relatedwork.tex
\section{Related work}

Although the first lane detection approaches rely on classical computer vision, substantial progress on accuracy and efficiency has been achieved with recent deep learning methods. Thus, this literature review focuses on deep lane detectors. This section first discusses the dominant approaches, which are based on segmentation~\cite{scnn,enet-sad,unetconvlstm,sim-cyclegan} or row-wise classification~\cite{intrakd,ufsa,e2e-lmd}, and, subsequently, review solutions in other directions. Finally, the lack of reproducibility (a common issue in lane detection works) is discussed.

\paragraph{Segmentation-based methods.}
In this approach, predictions are made on a per-pixel basis, classifying each pixel as either lane or background. With the segmentation map generated, a post-processing step is necessary to decode it into a set of lanes. In SCNN~\cite{scnn}, the authors propose a scheme specifically designed for long thin structures and show its effectiveness in lane detection. However, the method is slow (7.5 FPS), which hinders its applicability in real-world cases. Since larger backbones are one of the main culprits for slower speeds, the authors, in \cite{enet-sad}, proposes a self attention distillation (SAD) module to aggregate contextual information. The module allows the use of a more lightweight backbone, achieving a high-performance while maintaining real-time efficiency. In CurveLanes-NAS~\cite{curvelane-nas}, the authors propose the use of neural architecture search (NAS) to find a better backbone. Although they achieved state-of-the-art results, their NAS is extremely expensive computationally, requiring 5{,}000 GPU hours per dataset.

\paragraph{Row-wise classification methods.}
The row-wise classification approach is a simple way to detect lanes based on a grid division of the input image. For each row, the model predicts the most probable cell to contain a part of a lane marking. Since only one cell is selected on each row, this process is repeated for each possible lane in an image. Similar to segmentation methods, it also requires a post-processing step to construct the set of lanes. The method was first introduced in E2E-LMD~\cite{e2e-lmd}, achieving state-of-the-art results on two datasets. In~\cite{ufsa}, the authors show that it is capable of reaching high speed, although some accuracy is lost. This approach is also used in IntRA-KD~\cite{intrakd}.%

\paragraph{Other approaches.}
In FastDraw~\cite{fastdraw}, the author proposes a novel learning-based approach to decode the lane structures, which avoids the need for clustering post-processing steps (required in segmentation and row-wise classification methods). Although the proposed method is shown to achieve high speeds, it does not perform better than existing state-of-the-art methods in terms of accuracy. The same effect is shown in PolyLaneNet~\cite{polylanenet}, where an even faster model, based on deep polynomial regression, is proposed. In that approach, the model learns to output a polynomial for each lane. Despite its speed, the model struggles with the imbalanced nature of lane detection datasets, as evidenced by the high bias towards straight lanes in its predictions. In Line-CNN~\cite{linecnn}, an anchor-based method for lane detection is presented. This model achieves state-of-the-art results on a public dataset and promising results on another that is not publicly available. Despite the real-time efficiency, the model is considerably slower than other approaches. Moreover, the code is not public, which makes the results difficult to reproduce. There are also works addressing other parts of the pipeline of a lane detector. In~\cite{rgconstraints}, a post-processing method with a focus on occlusion cases is proposed, achieving results considerably higher than other works, but at the cost of notably low speeds (around 4 FPS). %

\paragraph{Reproducibility.}
As noted in~\cite{polylanenet}, many of the cited works do not publish the code to reproduce the results reported~\cite{linecnn,fastdraw,e2e-lmd}, or, in some cases, the code is only partially public~\cite{enet-sad,intrakd}. This hinders deeper qualitative and quantitative comparisons. For instance, the two most common metrics to measure a model's efficiency are multiply–accumulate operations (MACs) and frames-per-second (FPS). While the first does not depend on the benchmark platform, it is not always a good proxy for the second, which is the true goal. Therefore, FPS comparisons are also hindered by the lack of source code.

Unlike most of the previously proposed methods that managed to achieve high speeds at the cost of accuracy, we propose a method that is both faster and more accurate than existing state-of-the-art ones. In addition, the full code to reproduce the reported results is published for the community.

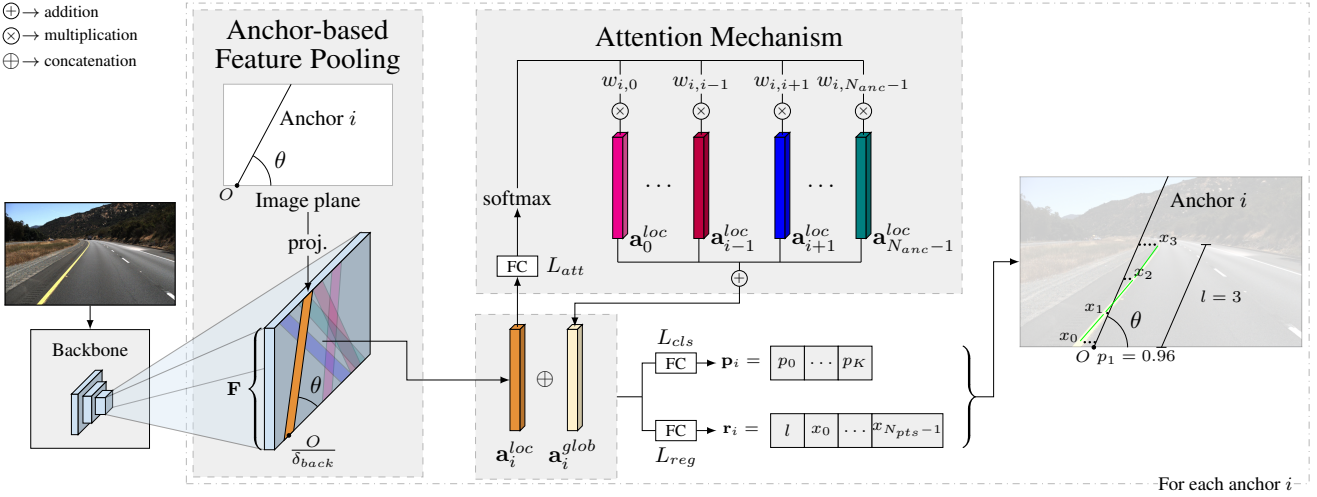
\begin{figure*}[t]
 	\centering
 	\resizebox{\textwidth}{!}{\input{figures/method_overview-v2}}
	\caption{Overview of the proposed method. A backbone generates feature maps from an input image. Subsequently, each anchor is projected onto the feature maps. This projection is used to pool features that are concatenated with another set of features created in the attention module. Finally, using this resulting feature set, two layers, one for classification and another for regression, make the final predictions.}
	\label{fig:overview}
\end{figure*}

%% file: figures/method_overview-v2.tex
\begin{tikzpicture}[every node/.style={inner sep=0,outer sep=0},x={(1,0)},y={(0,1)},z={({cos(45)},{sin(45)})}]
	
	\definecolor{grid_orange}{RGB}{230,145,56}
	\definecolor{grid_blue}{RGB}{207,226,243}
	\definecolor{vector_yellow}{RGB}{255,242,204}
	\definecolor{light_grey}{RGB}{239,239,239}
	\definecolor{verylight_grey}{RGB}{244,244,244}
	\definecolor{dark_grey}{RGB}{183,183,183}
	\definecolor{offset_orange}{RGB}{239,239,239}
	\definecolor{offset_darkorange}{RGB}{0,0,0}
	
	\newcommand\concat{\mathbin{+\mkern-10mu+}}
	
	\pgfdeclarelayer{z1}
	\pgfdeclarelayer{z2}
	\pgfdeclarelayer{z3}
	\pgfdeclarelayer{z4}
	\pgfsetlayers{background,z4,z3,z2,z1,main} %

	\tikzset{pics/conv/.style args={%
			#1 with dimensions #2 and #3 and #4}{
			code={
				\draw[ultra thin,fill=#1]  (0,0,0) coordinate(-front-bottom-left) to ++ (0,#3,0) coordinate(-front-top-left) --++ (#2,0,0) coordinate(-front-top-right) --++ (0,-#3,0)  coordinate(-front-bottom-right) -- cycle;
				\draw[ultra thin,fill=#1] (0,#3,0)  --++ (0,0,#4) coordinate(-back-top-left) --++ (#2,0,0)  coordinate(-back-top-right) --++ (0,0,-#4)  -- cycle;
				\draw[ultra thin,fill=#1!80!black] (#2,0,0) --++ (0,0,#4) coordinate(-back-bottom-right) --++ (0,#3,0) --++ (0,0,-#4) -- cycle;
				\coordinate (-right-bottom-left) at (#2, 0, 0);
				\coordinate (-right-bottom-right) at (#2, 0, #4);
				\coordinate (-right-top-left) at (#2, #3, 0);
				\coordinate (-right-top-right) at (#2, #3, #4);
				\coordinate (-right-center-bottom) at (#2,0,#4/2);
				\coordinate (-right-center-left) at (#2,#3/2,0);
				\coordinate (-right-center-right) at (#2,#3/2,#4);
				\coordinate (-right-center) at (#2,#3/2,#4/2);
				\coordinate (-front-left) at (0,#3/2,0);
				\coordinate (-front-right) at (#2,#3/2,0);
				\coordinate (-top-center) at (#2/2,#3,#4/2);
				\coordinate (-bottom-center) at (#2/2,0,0);
				\coordinate (-back-top-left) at (0, #3, #4);
				\coordinate (-back-bottom-left) at (0, #2, #4);
			}
	}}
	
	\node (input_img) at (0,0)%
	{%
		\setlength{\fboxsep}{0pt}%
		\setlength{\fboxrule}{0.5px}%
		\fbox{{\includegraphics[width=2.5cm, height=1.5cm]{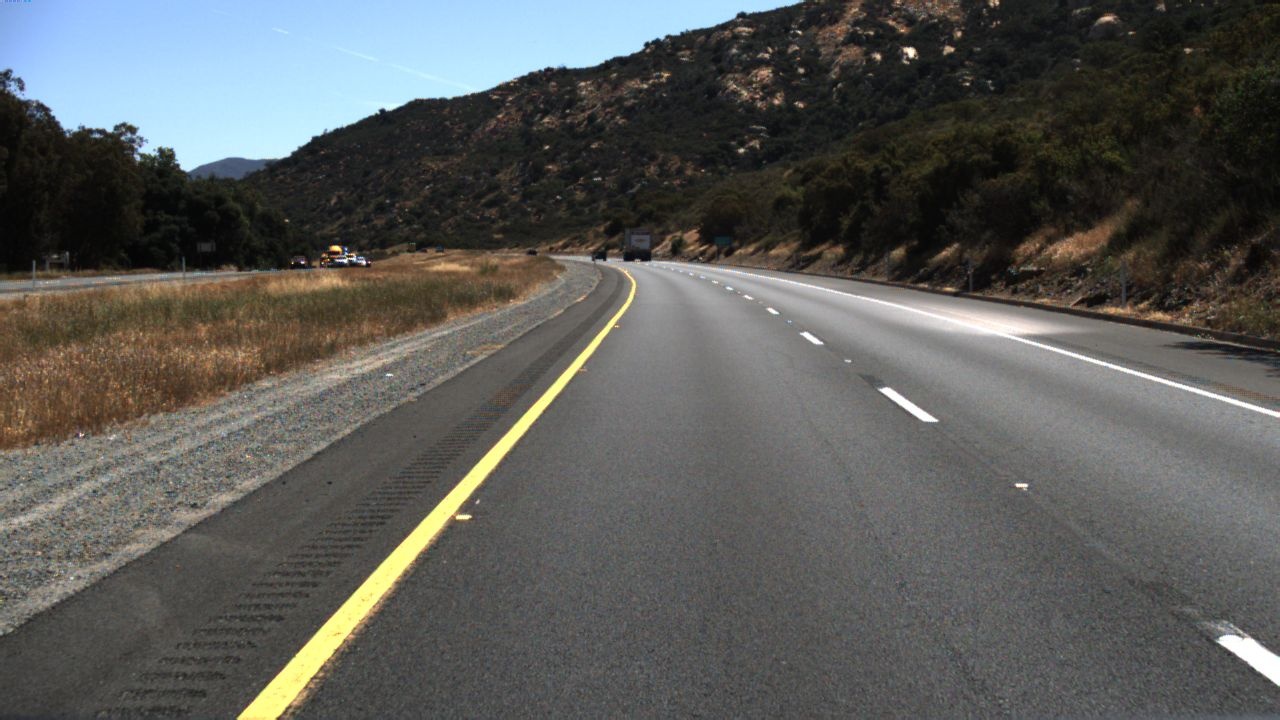}}}%
	};
	\begin{pgfonlayer}{z3}
		\node [rectangle, draw, fill=light_grey,node distance=2cm,
		text width=5em, text centered, minimum height=5em, below of=input_img] (backbone) {};
		\node[below=0.2cm] at (backbone.north) {\footnotesize Backbone};
		\foreach \X/\W/\H/\D [count=\i] in {-0.05/0.1/0.6/0.5,0/0.13/0.4/0.3,0.065/0.15/0.25/0.15} {
			\draw [scale=2] pic (box\i) at ($ (backbone.center) + (\X,-\H/4-0.1,-\D/4) $)
			{conv=grid_blue with dimensions {\W} and {\H} and {\D}};
		}  
	\end{pgfonlayer}
	\draw [-latex] (input_img.south) -- (backbone) node [midway] (bb_arrow) {};
	
	\node (fmap_start) at (backbone.south east) {};
	\draw [scale=2] pic (fmap) at ($(backbone.east) + (0.85, -0.45, 0)$) {conv=grid_blue with dimensions {0.15} and {1.8} and {2.0}};
	
	\coordinate (orange-strip-bl) at ($(fmap-right-bottom-left)!0.5!(fmap-right-bottom-right)$);
	\coordinate (orange-strip-br) at ($(fmap-right-bottom-left)!0.6!(fmap-right-bottom-right)$);
	\coordinate (orange-strip-tl) at ($(fmap-right-top-left)!0.6!(fmap-right-top-right)$);
	\coordinate (orange-strip-tr) at ($(fmap-right-top-left)!0.7!(fmap-right-top-right)$);
	\draw[ultra thin,fill=magenta, opacity=0.2] (orange-strip-bl) -- (orange-strip-br) -- (orange-strip-tr) -- (orange-strip-tl) -- cycle;
	
	\coordinate (orange-strip-bl) at ($(fmap-right-bottom-left)!0.8!(fmap-right-bottom-right)$);
	\coordinate (orange-strip-br) at ($(fmap-right-bottom-left)!0.9!(fmap-right-bottom-right)$);
	\coordinate (orange-strip-tl) at ($(fmap-right-top-left)!0.4!(fmap-right-top-right)$);
	\coordinate (orange-strip-tr) at ($(fmap-right-top-left)!0.5!(fmap-right-top-right)$);
	\draw[ultra thin,fill=purple, opacity=0.2] (orange-strip-bl) -- (orange-strip-br) -- (orange-strip-tr) -- (orange-strip-tl) -- cycle;
	
	\coordinate (orange-strip-bl) at ($(fmap-right-bottom-left)!0.7!(fmap-right-bottom-right)$);
	\coordinate (orange-strip-br) at ($(fmap-right-bottom-left)!0.8!(fmap-right-bottom-right)$);
	\coordinate (orange-strip-tl) at ($(fmap-right-top-left)!0.05!(fmap-right-top-right)$);
	\coordinate (orange-strip-tr) at ($(fmap-right-top-left)!0.15!(fmap-right-top-right)$);
	\draw[ultra thin,fill=blue, opacity=0.2] (orange-strip-bl) -- (orange-strip-br) -- (orange-strip-tr) -- (orange-strip-tl) -- cycle;
	
	\coordinate (orange-strip-bl) at ($(fmap-right-bottom-left)!0.85!(fmap-right-bottom-right)$);
	\coordinate (orange-strip-br) at ($(fmap-right-bottom-left)!0.95!(fmap-right-bottom-right)$);
	\coordinate (orange-strip-tl) at ($(fmap-right-top-left)!0.25!(fmap-right-top-right)$);
	\coordinate (orange-strip-tr) at ($(fmap-right-top-left)!0.35!(fmap-right-top-right)$);
	\draw[ultra thin,fill=teal, opacity=0.2] (orange-strip-bl) -- (orange-strip-br) -- (orange-strip-tr) -- (orange-strip-tl) -- cycle;

	\coordinate (orange-strip-bl) at ($(fmap-right-bottom-left)!0.1!(fmap-right-bottom-right)$);
	\coordinate (orange-strip-br) at ($(fmap-right-bottom-left)!0.2!(fmap-right-bottom-right)$);
	\coordinate (orange-strip-tl) at ($(fmap-right-top-left)!0.3!(fmap-right-top-right)$);
	\coordinate (orange-strip-tr) at ($(fmap-right-top-left)!0.4!(fmap-right-top-right)$);
	\draw[ultra thin,fill=grid_orange] (orange-strip-bl) -- (orange-strip-br) -- (orange-strip-tr) -- (orange-strip-tl) -- cycle;
	
	\coordinate (orange-strip-bm) at ($(orange-strip-br)!0.5!(orange-strip-bl)$);
	\node [below right] (strip-org-label) at ($(orange-strip-bm) + (0, 0)$) {$\frac{O}{\delta_{back}}$};
	\pic [draw, "$\theta$", angle eccentricity=1.5] {angle = fmap-right-bottom-right--orange-strip-br--orange-strip-tr};
	\filldraw (orange-strip-bm) circle (0.75pt);
	
	\draw [draw, fill=grid_blue, ultra thin, fill, opacity=0.2] (box3-front-top-right) -- (fmap-front-top-left) -- (fmap-front-bottom-left) -- (box3-front-bottom-right);
	
	\begin{pgfonlayer}{z2}
		\draw [draw, fill=grid_blue, ultra thin, fill, opacity=0.2] (box3-right-bottom-right) -- (fmap-back-bottom-left) -- (fmap-front-bottom-left) --
		(box3-front-bottom-right);
		\draw [draw, fill=grid_blue, ultra thin, fill, opacity=0.2] (box3-back-top-right) -- (fmap-back-top-left) -- (fmap-back-bottom-left) --
		(box3-back-bottom-right);
		\draw [draw, fill=grid_blue, ultra thin, fill, opacity=0.2] (box3-front-top-right) -- (fmap-front-top-left) -- (fmap-back-top-left) -- (box3-back-top-right);
	\end{pgfonlayer}
	
	\draw [decoration={calligraphic brace,amplitude=5pt}, decorate, line width=1pt]
	($ (fmap-front-bottom-left) + (-0.05,0,0) $) -- ($ (fmap-front-top-left) + (-0.05,0,0) $) node [black,midway,xshift=-0.4cm]
	{\footnotesize $\mathbf{F}$};
	
	\coordinate (orange-strip-tm) at ($(orange-strip-tr)!0.5!(orange-strip-tl)$);
	\draw [shorten >= 1pt, shorten <= 1pt, latex-] (orange-strip-tm) to ($(orange-strip-tm) + (0, 0.6125)$) node[above]{\small proj.};
	\draw [shorten <= 8pt] ($(orange-strip-tm) + (0, 0.6125)$) to ($(orange-strip-tm) + (0, 1.225)$) coordinate(strip-arrow);

	\node [above] (image_plane_label) at (strip-arrow) {\small Image plane};
	\node [above=0.25cm, draw, dark_grey, fill=white, minimum width=2.5cm, minimum height=1.5cm] (image_plane) at (image_plane_label) {};
	\node [above=0.15cm, text width=3cm, align=center] (pooling_title) at (image_plane.north) {\large Anchor-based Feature Pooling};
	\draw [thin] ($(image_plane.south west) + (0.2,0)$) coordinate(anchor-line-origin) -- ($(image_plane.north west) + (1.0,0)$) coordinate(anchor-line);
	\filldraw (anchor-line-origin) circle (0.75pt);
	\node at ($(anchor-line) + (0.4, -0.5)$) {\small Anchor $i$};
	\node [below left]  at ($(anchor-line-origin) + (-0.05, -0.05)$) {\scriptsize $O$};
	\coordinate (x-axis) at (image_plane.south);
	\pic [draw, "$\theta$", angle eccentricity=1.5] {angle = x-axis--anchor-line-origin--anchor-line};

	\begin{pgfonlayer}{z3}
		\node [draw, dark_grey, dashed, fill=light_grey, inner ysep=0.15cm, inner xsep=0.2cm, fit={(pooling_title) (image_plane) (strip-org-label) (fmap-front-bottom-left)}] (pooling_bg) {};
	\end{pgfonlayer}

	\draw [scale=2] pic (anchor_fmap) at ($(fmap-right-center-bottom) + (1.6, 0, -0.3)$) {conv=grid_orange with dimensions {0.15} and {1.5} and {0.1}};
	\node [below of=anchor_fmap-bottom-center, node distance=0.35cm] (anchor_fmap_label) {$\mathbf{a}_i^{loc}$};
	\draw [scale=2] pic (att_fmap) at ($(anchor_fmap-front-bottom-right) + (0.35, 0, 0)$) {conv=vector_yellow with dimensions {0.15} and {1.5} and {0.1}};
	\node [below of=att_fmap-bottom-center, node distance=0.35cm] (att_fmap_label) {$\mathbf{a}_i^{glob}$};
	\node (concat_symbol) at ($(anchor_fmap-right-center)!0.5!(att_fmap-front-left)$) {$\oplus$};
	\begin{pgfonlayer}{z2}
		\node [draw, dark_grey, dashed, fill=light_grey, inner ysep=0.15cm, inner xsep=0.3cm, fit={(anchor_fmap-back-top-left) (att_fmap-front-bottom-right) (att_fmap_label) (anchor_fmap_label)}] (concat_fmap_bg) {};
	\end{pgfonlayer}
	\draw [-latex] (fmap-right-center) --++ (1.25, 0) |- (anchor_fmap-front-left);
	
	\node [draw, fill=white, shape=rectangle, minimum width=0.6cm, minimum height=0.3cm, anchor=center] (latt_layer) at ($(anchor_fmap-top-center) + (0, 0.9, 0)$) {\scriptsize FC};
	\node [right=0.125cm of latt_layer] (latt_text) {\small $L_{att}$};
	\draw [-latex] (anchor_fmap-top-center) to (latt_layer.south);
	\node [above= 0.75cm of latt_layer] (softmax) {\small softmax};
	\draw [shorten >= 1pt,-latex] (latt_layer.north) to (softmax.south);
	
	\foreach \X/\fmapcolor/\w [count=\i] in {0/magenta/0,0.6/purple/i-1,1.2/blue/i+1,1.8/teal/N_{anc}-1} {
		\draw [scale=2] pic (a_\i) at ($(softmax.east) + (0.45 + \X, -0.3, 0)$) {conv={\fmapcolor} with dimensions {0.15} and {1.5} and {0.1}};
		\node [draw, shape=circle, minimum width=0.2cm, minimum height=0.2cm, anchor=center] (a_\i_times)at ($(a_\i-top-center) + (0, 0.35, 0)$) {\tiny $\times$};
		\node [align=center, anchor=center] (a_\i_w) at ($(a_\i-top-center) + (0, 0.75, 0)$) {\small $w_{i,\w}$};
		
		\draw (a_\i-top-center) to (a_\i_times);
		\draw [shorten >= 1pt] (a_\i_times) to (a_\i_w);
		\draw [shorten <= 1pt] (a_\i_w) to ($(a_\i_w) + (0,0.35)$);
		\draw (a_\i-bottom-center) to ($(a_\i-bottom-center) + (0, -0.35)$) coordinate(a_\i-line-bottom);
		\node [right=0.1cm] (a_\i-label) at (a_\i-front-bottom-right) {$\mathbf{a}_{\w}^{loc}$};
	}
	
	\draw [shorten <= 1pt] (softmax) |- ($(a_4_w) + (0,0.35)$);
	\node [anchor=center] at ($(a_1-right-center-right)!0.5!(a_2-front-left)$) {$\dots$};
	\node [anchor=center] at ($(a_3-right-center-right)!0.5!(a_4-front-left)$) {$\dots$};
	
	\draw (a_1-line-bottom) to (a_4-line-bottom);
	\draw ($(a_2-line-bottom)!0.5!(a_3-line-bottom)$) --++ (0, -0.125) coordinate(sum-top);
	\node [draw, shape=circle, minimum width=0.2cm, minimum height=0.2cm, anchor=north] (sum_symbol) at (sum-top) {\tiny $+$};
	\draw [-latex] (sum_symbol.south) --++ (0, -0.25) -| (att_fmap-top-center);
	
	\begin{pgfonlayer}{z2}
		\node [draw, dark_grey, dashed, fill=light_grey, inner ysep=0.6cm, inner xsep=0.1cm, fit={(softmax) ($(a_4_w) + (0, 0.5)$) (a_4-label)}] (att_mech_bg) {};
		\node [below=0.25cm] at (att_mech_bg.north) {\large Attention Mechanism};
	\end{pgfonlayer}
	
	\draw (concat_fmap_bg.east) --++ (0.375, 0) coordinate(output_fork);
	\draw (output_fork) |-++ (0.2, 0.5) coordinate(cls_line);
	\node [draw, shape=rectangle, minimum width=0.6cm, minimum height=0.3cm, anchor=west] (lcls_layer) at (cls_line) {\scriptsize FC};
	\node [above=0.1cm] at (lcls_layer.north) {\small $L_{cls}$};
	\draw [-latex] (lcls_layer.east) --++ (0.3,0) coordinate(cls_arrow);
	\node [anchor=west, minimum width=0.8cm] (pi) at (cls_arrow) {\scriptsize $~\mathbf{p}_i=~$};
	\foreach \i/\name in {0/$p_0$,1/$\ldots$,2/$p_K$} {
		\pgfmathsetmacro\x{0.5*\i}
		\node [rectangle, draw, minimum width=0.5cm, minimum height=0.5cm, fill=light_grey, text centered, inner sep=0.125, anchor=west] (cls\i) at ($(pi.east) + (\x, 0)$) {\scriptsize \name};
	}
	\draw (output_fork) |-++ (0.2, -0.5) coordinate(reg_line);
	\node [draw, shape=rectangle, minimum width=0.6cm, minimum height=0.3cm, anchor=west] (lreg_layer) at (reg_line) {\scriptsize FC};
	\node [below=0.1cm] at (lreg_layer.south) {\small $L_{reg}$};
	\draw [-latex] (lreg_layer.east) --++ (0.3,0) coordinate(reg_arrow);
	\node [anchor=west, minimum width=0.8cm] (ri) at (reg_arrow) {\scriptsize $~\mathbf{r}_i=~$};
	\foreach \i/\name/\cor in {0/$l$/light_grey,1/$x_0$/offset_orange,2/$\ldots$/offset_orange,3/$x_{N_{pts}-1}$/offset_orange} {
		\pgfmathsetmacro\x{0.5*\i}
		\node [rectangle, draw, minimum width=0.5cm, minimum height=0.5cm, fill=\cor, text centered, inner sep=0.125, anchor=west] (reg\i) at ($(ri.east) + (\x, 0)$) {\scriptsize \name};
	}
	
	\node (outputs_intersection) at (reg3.south east |- cls2.north east) {};
	\draw [decoration={mirror, calligraphic brace,amplitude=5pt}, decorate, line width=1pt]
	($ (reg3.south east) + (0.3,0,0) $) -- ($ (outputs_intersection) + (0.3,0,0) $) node [black,midway,xshift=0.15cm]
	(output_brace_center){};
	
	\node [opacity=0.2](output_img) at ($(reg3) + (3.75,2.5)$)%
	{%
		\setlength{\fboxsep}{0pt}%
		\setlength{\fboxrule}{0.5px}%
		\fbox{{\includegraphics[width=4.15cm, height=2.5cm]{3.jpg}}}%
	};
	
	\draw [thin] ($(output_img.south west) + (1.1,0)$) coordinate(anchor-line-origin) -- ($(output_img.north west) + (2.2,0)$) coordinate(anchor-line) node [pos=0.03](offset1) {} node [pos=0.2](offset2) {} node [pos=0.4](offset3) {} node [pos=0.6](offset4) {};
	\filldraw (anchor-line-origin) circle (0.75pt);
	\node at ($(anchor-line) + (0.6, -0.35)$) {\small Anchor $i$};
	\node [below left]  at ($(anchor-line-origin) + (-0.05, -0.05)$) {\scriptsize $O$};
	\coordinate (x-axis) at (output_img.south);
	\draw [-latex] (output_brace_center) --++ (0.3, 0) |- (output_img.west);
	\pic [draw, "$\theta$", angle eccentricity=1.5] {angle = x-axis--anchor-line-origin--anchor-line};
	\node [below right]  at ($(anchor-line-origin) + (0.05, -0.05)$) {\scriptsize $p_1=0.96$};
	\node (pred1) at ($(offset1) + (-0.25, 0)$) {};
	\node (pred2) at ($(offset2) + (-0.06, 0)$) {};
	\node (pred3) at ($(offset3) + (0.15, 0)$) {};
	\node (pred4) at ($(offset4) + (0.3, 0)$) {};
	\draw [green] (pred1) -- (pred2) -- (pred3) -- (pred4);
	\draw [line width=1pt, line cap=round, dash pattern=on 0pt off 2\pgflinewidth] (offset1) to (pred1) node[above left, offset_darkorange] {\scriptsize $x_0$};
	\draw [line width=1pt, line cap=round, dash pattern=on 0pt off 2\pgflinewidth] (offset2) to (pred2) node[above left,offset_darkorange] {\scriptsize $x_1$};
	\draw [line width=1pt, line cap=round, dash pattern=on 0pt off 2\pgflinewidth] (offset3) to (pred3) node[above right,offset_darkorange] {\scriptsize $x_2$};
	\draw [line width=1pt, line cap=round, dash pattern=on 0pt off 2\pgflinewidth](offset4) to (pred4) node[above right,offset_darkorange] {\scriptsize $x_3$};
	\draw [thin] ($(anchor-line-origin) + (1,0)$) -- ($(offset4) + (1,0)$) node[midway, right=0.25cm] {\scriptsize $l=3$};
	\draw [thin] ($(offset4) + (0.925,0.0)$) -- ($(offset4) + (1.075,0)$);
	\draw [thin] ($(anchor-line-origin) + (0.925,0)$) -- ($(anchor-line-origin) + (1.075,0)$);

	\begin{pgfonlayer}{z4}
		\node [draw, dark_grey, dashdotted, inner sep=0.1cm, fit={(pooling_bg) (att_mech_bg) (reg3) (output_img)}] (for_each_bg) {};
		\node [fill=white, left=0.25cm] at (for_each_bg.south east) {\footnotesize For each anchor $i$};
	\end{pgfonlayer}
	
	\node (top-left-corner) at (input_img.north west |- for_each_bg.north east) {};
	\node (legend) at ($(top-left-corner) + (0.1cm, 0)$) {};
	\node [below, draw, shape=circle, minimum width=0.2cm, minimum height=0.2cm] (sum_symbol_legend) at (legend) {\tiny $+$};
	\node [right] at ($(sum_symbol_legend) + (0.15cm, 0)$) {\scriptsize $\rightarrow$ addition};
	\node [draw, shape=circle, minimum width=0.2cm, minimum height=0.2cm, below=0.25cm] (mult_legend) at (sum_symbol_legend) {\tiny $\times$};
	\node [right] at ($(mult_legend) + (0.15cm, 0)$) {\scriptsize $\rightarrow$ multiplication};
	\node [below=0.25cm](concat_symbol_legend) at (mult_legend) {$\oplus$};
	\node [right] at ($(concat_symbol_legend) + (0.15cm, 0)$) {\scriptsize $\rightarrow$ concatenation};

\end{tikzpicture}

%% file: secs/3_method.tex
\section{Proposed method}

\methodname~is an anchor-based single-stage model (like YOLOv3~\cite{yolo} or SSD~\cite{ssd}) for lane detection. An overview of the method is shown in~\Cref{fig:overview}. It receives as input RGB images $I \in \mathbb{R}^{3 \times H_I \times W_I}$ taken from a front-facing camera mounted in a vehicle. The outputs are lane boundary lines (hereafter called lanes, following the usual terminology in the literature). To generate those outputs, a convolutional neural network (CNN), referred to as the backbone, generates a feature map that is then pooled to extract each anchor's features. Those features are combined with a set of global features produced by an attention module. By combining local and global features, the model can use information from other lanes more easily, which might be necessary in cases with conditions such as occlusion or no visible lane markings. Finally, the combined features are passed to fully-connected layers to predict the final output lanes.

\subsection{Lane and anchor representation}
\label{sec:rep}

A lane is represented by 2D-points with equally-spaced y-coordinates $Y = \{y_i\}_{i=0}^{N_{pts}-1}$, where $y_i= i \cdot \frac{H_I}{N_{pts} - 1}$. Since $Y$ is fixed, a lane can then be defined only by its x-coordinates $X = \{x_i\}_{i=0}^{N_{pts}-1}$, each $x_i$ associated with the respective $y_i \in Y$. Since most lanes do not cross the whole image vertically, a start-index $s$ and an end-index $e$ are used to define the valid contiguous sequence of $X$.

Likewise Line-CNN~\cite{linecnn}, our method performs anchor-based detection using lines instead of boxes, which means that lanes' proposals are made having these lines as references. An anchor is a ``virtual'' line in the image plane defined by (i) an origin point $O = (x_{orig}, y_{orig})$ (with $y_{orig} \in Y$) located in one of the borders of the image (except the top border) and (ii) a direction $\theta$. The proposed method uses the same set of anchors as~\cite{linecnn}. %
This lane and anchor representation satisfies the vast majority of real-world lanes.

\subsection{Backbone}

The first stage of the proposed method is feature extraction, which can be performed by any generic CNN, such as a ResNet~\cite{he2016deep}. The output of this stage is a feature map $\mathbf{F}_{back} \in \mathbb{R}^{C_F' \times H_F \times W_F}$ from which the features for each anchor will be extracted through a pooling process, as described in the next section. For dimensionality reduction, a $1 \times 1$ convolution is applied onto $\mathbf{F}_{back}$, generating a channel-wise reduced feature map $\mathbf{F} \in \mathbb{R}^{C_F \times H_F \times W_F}$. This reduction is performed to reduce computational costs.

\subsection{Anchor-based feature pooling}

An anchor defines the points of $\mathbf{F}$ that will be used for the respective proposals. Since the anchors are modeled as lines, the interest points for a given anchor are those that intercept the anchor's virtual line (considering the rasterized line reduced to the feature maps dimensions). For every $y_j = 0, 1, 2, \ldots, H_F - 1$, there will be a single corresponding x-coordinate,
\begin{equation}
x_j = \left\lfloor\frac{1}{\tan{\theta}}~(y_j - y_{orig}/\delta_{back}) + x_{orig}/\delta_{back}\right\rfloor,
\label{eq:proj}    
\end{equation}
where $(x_{orig}$, $y_{orig})$ and $\theta$ are, respectively, the origin point and slope of the anchor's line, and $\delta^{back}$ is the backbone's global stride. Thus, every anchor $i$ will have its corresponding feature vector $\mathbf{a}^{loc}_i \in \mathbb{R}^{C_F \cdot H_F}$ (column-vector notation) pooled from $\mathbf{F}$ that carries local feature information (local features). In cases where a part of the anchor is outside the boundaries of $\mathbf{F}$, $\mathbf{a}^{loc}_i$ is zero-padded.

Notice that the pooling operation is similar to the Fast R-CNN's~\cite{fastrcnn} region of interest projection (RoI projection), however, instead of using the proposal for pooling, a single-stage detector is achieved by using the anchor itself. Additionally, the RoI pooling layer (used to generate fixed-size features) is not necessary for our method. Comparing to Line-CNN~\cite{linecnn}, that leverages only the feature maps' borders, our method can potentially explore all the feature map, which enables the use of more lightweight backbones with smaller receptive field.

\subsection{Attention mechanism}

Depending on the model architecture, the information carried by the pooled feature vector ends up being mostly local. This is particularly the case for shallower and faster models, which tend to exploit backbones with smaller receptive fields. However, in some cases (such as the ones with occlusion) the local information may not be enough to predict the lane's existence and its position. To address that problem, we propose an attention mechanism that acts on the local features ($\mathbf{a}^{loc}_{\sbullet}$) to produce additional features ($\mathbf{a}^{glob}_{\sbullet}$) that aggregate global information.%

Basically, the attention mechanism structure is composed of a fully-connected layer $L_{att}$ which processes a local feature vector $\mathbf{a}^{loc}_{i}$ and outputs a probability (weight) $w_{i,j}$ for every anchor $j$, $j \neq i$. Formally,
\begin{equation}
    \label{eq:att}
    w_{i,j} = 
      \begin{cases}
        \operatorname{softmax}(L_{att}(\mathbf{a}^{loc}_i))_j, & \text{if } j < i\\
        0, & \text{if } j = i \\
        \operatorname{softmax}(L_{att}(\mathbf{a}^{loc}_i))_{j-1}, & \text{if } j > i
      \end{cases}
\end{equation}
Afterwards, those weights are combined with the local features to produce a global feature vector of same dimension:
\begin{equation}
	\mathbf{a}^{glob}_i = \sum_{j} w_{i,j}~\mathbf{a}^{loc}_j.
\end{equation}
Naturally, the whole process can be implemented efficiently with matrix multiplication, since the same procedure is executed for all anchors. Let $N_{anc}$ be the number of anchors. Let $\mathbf{A}^{loc} = [\mathbf{a}^{loc}_0, \ldots, \mathbf{a}^{loc}_{N_{anc} - 1}]^T$ be the matrix containing the local feature vectors (as rows) and $\mathbf{W}=[w_{i,j}]_{N_{anc} \times N_{anc}}$ the weight matrix, $w_{i,j}$ defined in \Cref{eq:att}. Thus, global features can be computed as:

\begin{equation}
	\mathbf{A}^{glob} = \mathbf{W}~\mathbf{A}^{loc}.
\end{equation}
Notice that $\mathbf{A}^{glob}$ and $\mathbf{A}^{loc}$ have the same dimensions, \ie, $\mathbf{A}^{glob} \in \mathbb{R}^{N_{anc} \times C_F\cdot H_F}$.

\subsection{Proposal prediction}
\label{sec:proposal_prediction}
A lane proposal is predicted for each anchor and consists of three main components: (i) $K + 1$ probabilities ($K$ lane types and one class for ``background'' or invalid proposal), (ii) $N_{pts}$ offsets (the horizontal distance between the prediction and the anchor's line), and (iii) the length $l$ of the proposal (the number of valid offsets). The start-index ($s$) for the proposal is directly determined by the y-coordinate of the anchor's origin ($y_{orig}$, see \Cref{sec:rep}). Thus, the end-index can be determined as $e = s + \lfloor l \rceil - 1$.

To generate the final proposals, local and global information are aggregated by concatenating $\mathbf{a}^{loc}_{i}$ and $\mathbf{a}^{glob}_{i}$, producing an augmented feature vector $\mathbf{a}^{aug}_{i} \in \mathbb{R}^{2 \cdot C_F\cdot H_F}$. This augmented vector is fed to two parallel fully-connected layers, one for classification ($L_{cls}$) and one for regression ($L_{reg}$), which produce the final proposals.
$L_{cls}$ predicts $\mathbf{p}_i = \{p_0, \ldots, p_{K+1}\}$ (item i) and $L_{reg}$ predicts $\mathbf{r}_i = \left(l, \{x_0, \ldots, x_{N_{pts} - 1}\}\right)$ (items ii and iii).

\subsection{Non-maximum Supression (NMS)}

As usual in anchor-based deep detection, NMS is paramount to reduce the number of false positives. In the proposed method, this procedure is applied both during training and test phases based on the lane distance metric proposed in~\cite{linecnn}. The distance between two lanes $X_a=\{x^a_i\}_{i=1}^{N_{pts}}$ and $X_b=\{x^b_i\}_{i=1}^{N_{pts}}$ is computed based on their common valid indices (or y-coordinates). Let $s'= \max(s_{a}, s_{b})$ and $e'=\min(e_{a}, e_{b})$ define the range of those common indices. Thus, the lane distance metric is defined as
\begin{equation}
	D(X_a, X_b) =
	\begin{cases}
		\frac{1}{e' -s' + 1} \cdot \sum_{i=s'}^{e'} | x_{i}^{a} - x_{i}^{b} |, & e' \ge s' \\
		+\infty, & \text{ otherwise.}
	\end{cases}
	\label{eq:line-distance}
\end{equation}

\subsection{Model training}

During training, the distance metric in \Cref{eq:line-distance} is also used to define the positive and the negative anchors. First, the metric is used to measure the distance between every anchor (those not filtered in NMS) and the ground-truth lanes. Subsequently, the anchors with distance (Eq. \ref{eq:line-distance}) lower than a threshold $\tau_p$ are considered positives, while those with distance greater than $\tau_n$ are considered negatives. Anchors (and their associated proposals) with distance in between those thresholds are disregarded. The remainder $N_{p\&n}$ are used in a multi-task loss defined as:

\begin{equation}
\begin{aligned}
	\mathcal{L}(\{\mathbf{p}_i, \mathbf{r}_i\}_{i=0}^{N_{p\&n}-1}) = &~\lambda\:\sum_{i} \mathcal{L}_{cls}(\mathbf{p}_i, \mathbf{p}_i^*) \\
	& + \sum_{i}\mathcal{L}_{reg}(\mathbf{r}_i, \mathbf{r}_i^*),
\end{aligned}
\end{equation}
where $\mathbf{p}_i$, $\mathbf{r}_i$ are the classification and regression outputs for the anchor $i$, whereas $\mathbf{p}_i^*$ and $\mathbf{r}_i^*$ are the classification and regression targets for the anchor $i$. The regression loss is computed only with the length $l$ and the x-coordinates values corresponding to indices common to both the proposal and the ground-truth. The common indices (between $s'$ and $e'$) of the x-coordinates are selected similarly to the lane distance (\Cref{eq:line-distance}) but with $e'=e_{gt}$ instead of $e'=\min(e_{prop}, e_{gt})$, where $e_{prop}$ and $e_{gt}$ are the end-indexes for the proposal and its associated ground-truth, respectively. If the end-index predicted in the proposal $e_{prop}$ is used, the training may become unstable by converging to degenerate solutions (\eg, $e_{prop}$ might converge to zero). The functions $\mathcal{L}_{cls}$ and $\mathcal{L}_{reg}$ are the Focal Loss~\cite{focal-loss} and the Smooth L1, respectively. If the anchor $i$ is considered negative, its corresponding $\mathcal{L}_{reg}$ is equal to $0$. The factor $\lambda$ is used to balance the loss components.

\subsection{Anchor filtering for speed efficiency}
\label{met:anchor-filtering}

The full set of anchors comprises a total of 2{,}782 anchors. This elevated number is one of the main factors limiting the model's speed. Since a large number of anchors will not be useful during the training (\eg, some anchors may have a starting point above the horizon line of all images in the training dataset), the set's size can be reduced. To choose which anchors are going to be disregarded in both training and test phases, the method measures the number of times each anchor from the training set is marked as positive (same criteria as in the training). Finally, only the top-$N_{anc}$ marked anchors are kept for further processing (also during test).

%% file: secs/4_experiments.tex
\section{Experiments}
\begin{figure}
	\centering
	\subfigure{
	    \begin{minipage}[t]{0.48\textwidth}
    		\centering
    		\includegraphics[width=0.32\columnwidth]{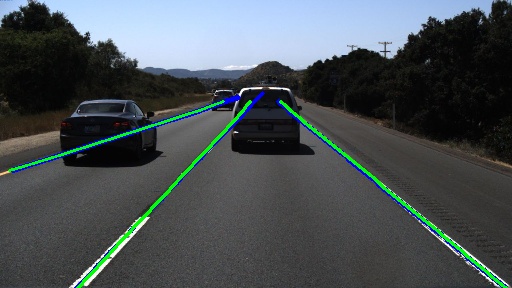}
    		\includegraphics[width=0.32\columnwidth]{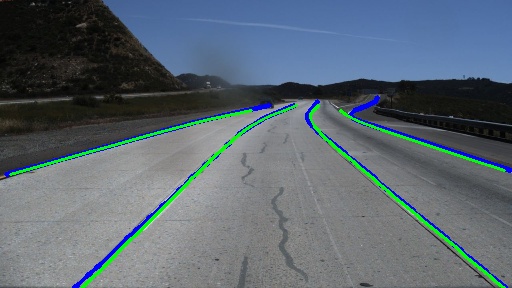}
    		\includegraphics[width=0.32\columnwidth]{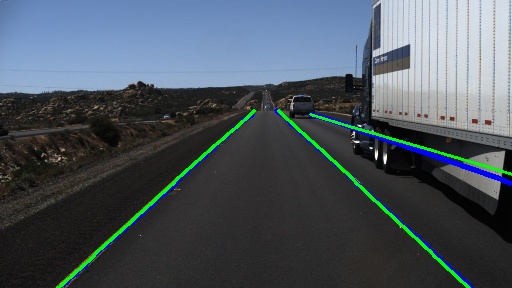}
        \end{minipage}
	}
	\vskip -2pt
	\subfigure{
	    \begin{minipage}[t]{0.48\textwidth}
    		\centering
    		\includegraphics[width=0.32\columnwidth]{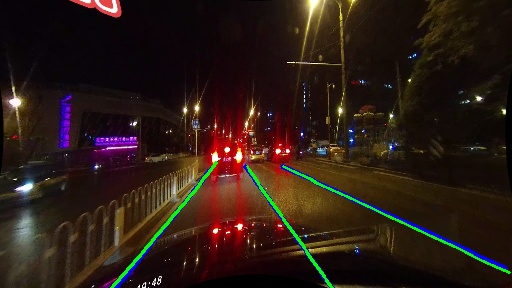}
    		\includegraphics[width=0.32\columnwidth]{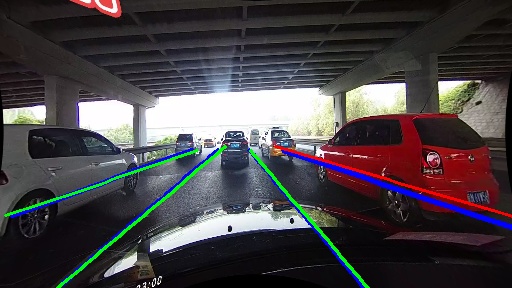}
    		\includegraphics[width=0.32\columnwidth]{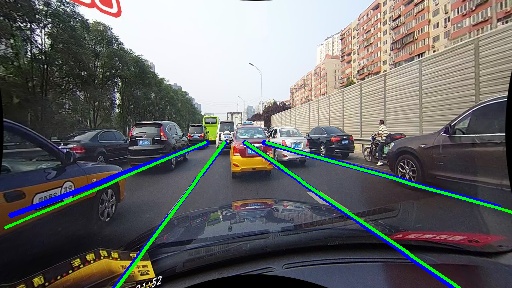}
        \end{minipage}
	}
	\vskip -2pt
	\subfigure{
	    \begin{minipage}[t]{0.48\textwidth}
    		\centering
    		\includegraphics[width=0.32\columnwidth]{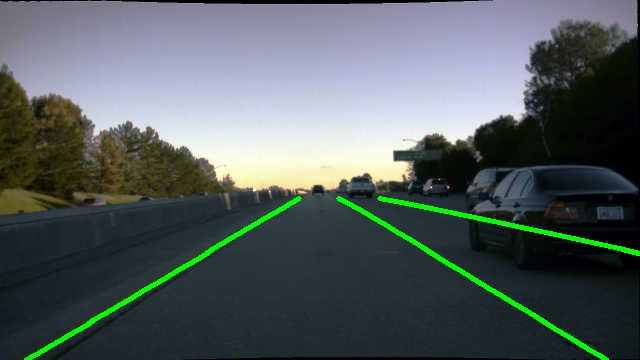}
    		\includegraphics[width=0.32\columnwidth]{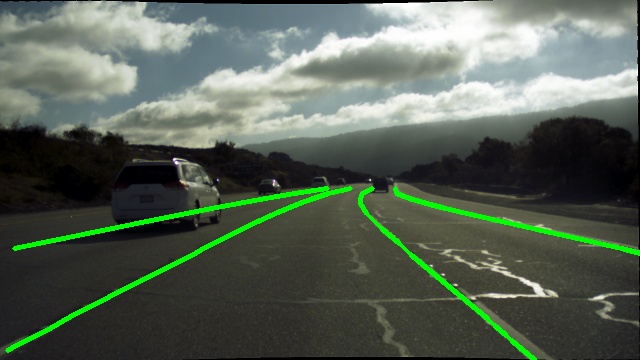}
    		\includegraphics[width=0.32\columnwidth]{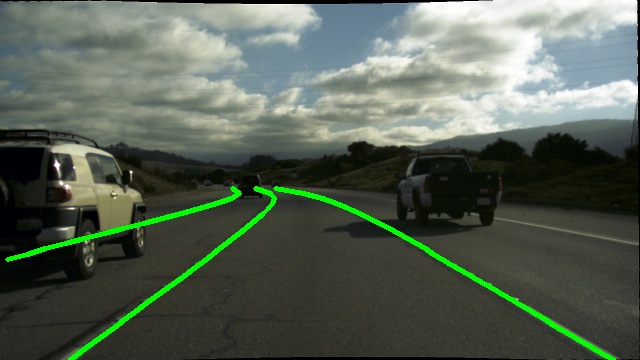}
        \end{minipage}
	}
	\vskip 4pt
    \centering
	\caption{\methodname~qualitative results on TuSimple (top row), CULane (middle row), and LLAMAS (bottom row). Blue lines are ground-truth, while green and red lines are true-positives and false-positives, respectively.}
	\label{fig:qualitative}
\end{figure}

\begin{figure*}[t]
	\begin{center}
	\resizebox{\textwidth}{!}{\input{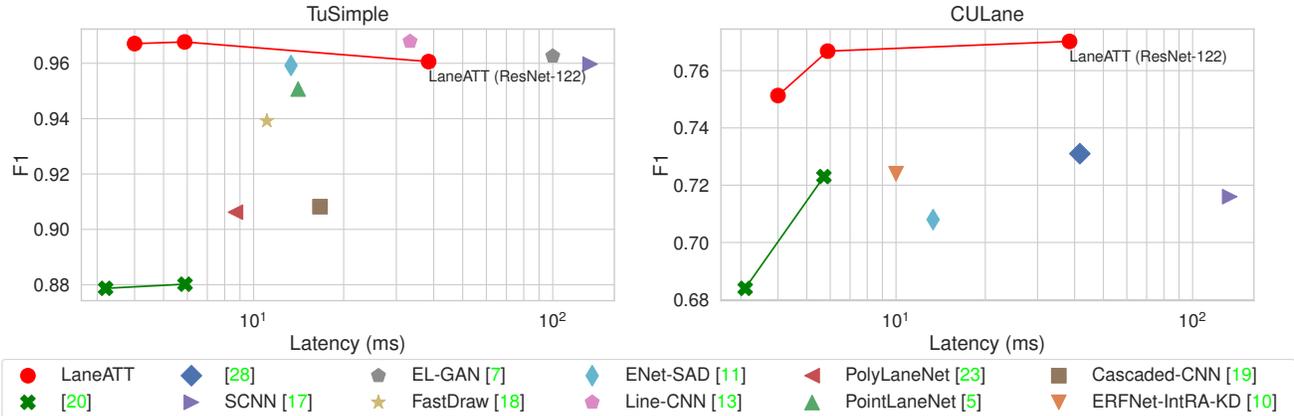}}
	\end{center}
	\vspace{-12pt}
	\caption{Model latency vs. F1 of state-of-the-art methods on CULane and TuSimple.}
	\label{fig:fps-vs-acc}
\end{figure*}

Our method was evaluated on the two most widely used lane detection datasets (TuSimple~\cite{tusimple} and CULane~\cite{scnn}) and on the recently released benchmark (LLAMAS~\cite{llamas}). An overview of the datasets can be seen in~\Cref{tab:datasets}. This section starts describing the efficiency metrics and some of the implementation details. All experiments used the default metric parameters set by the dataset's creator, which are the same used by the related works. The three first subsections discusses the experimental results for each dataset (including the dataset description and the evaluation metrics). The two final subsections address experiments on efficiency trade-offs and an ablation study on parts.

\input{tables/datasets}

\paragraph{Efficiency metrics.}
Two efficiency-related metrics are reported: frames-per-second (FPS) and multiply-accumulate operations (MACs). One MAC is approx. two floating operations (FLOPs). The FPS is computed using a single image per batch and constant inputs, so the metric is not dependent on I/O operations but only on the model's efficiency.

\paragraph{Implementation details.}
Except when explicitly indicated, all input images are resized to $H_I \times W_I = 360 \times 640$ pixels. For all training sessions, the Adam optimizer is used for 15 epochs on CULane and 100 epochs on TuSimple (the large discrepancy is due to the large difference between the datasets' sizes). For data augmentation, a random affine transformation is performed (with translation, rotation, and scaling) along with random horizontal flips. Most experiments and all FPS measures were computed on a machine with an Intel i9-9900KS and an RTX 2080 Ti. The model parameters were $N_{pts}=72$, $N_{anc}=1000$, $\tau_{p}=15$ and $\tau_n=20$. The used datasets do not provide the lane type (\eg, dashed or solid), thus, we set $K=1$ (according \Cref{sec:proposal_prediction}). For more details and parameter values, the code\footnote{https://github.com/lucastabelini/LaneATT} can be accessed, along with each experiment's configuration.

\subsection{TuSimple}
\paragraph{Dataset description.}
TuSimple~\cite{tusimple} is a lane detection dataset containing only highway scenes, a scenario that is usually considered easier than street scenes. Despite that, it is one of the most widely used datasets in lane detection works. All images have $1280 \times 720$ pixels, with at most 5 lanes.

\input{tables/tusimple}
\input{tables/culane}

\paragraph{Evaluation metrics.}
On TuSimple, the three standard metrics are false discovery rate (FDR), false negative rate (FNR), and accuracy. The accuracy is defined as

\begin{equation}
	\text{Acc} = \frac{\sum_{clip} C_{clip}}{\sum_{clip} S_{clip}},
\end{equation}
where $C_{clip}$ is the number of lane points predicted correctly in the clip and $S_{clip}$ is the total number of points in the clip (or image). For a point prediction to be considered correct, the prediction has to be within 20 pixels the ground truth. For a lane prediction to be considered a true positive (for the FDR and FNR metrics), its number of correct predicted points has to be greater than 85\%. We also report the F1 score (hereafter called F1), which is the harmonic mean of the precision and the recall.

\paragraph{Results.}

The results of \methodname~on TuSimple, along with other state-of-the-art methods, are shown in~\Cref{tab:tusimple} and in~\Cref{fig:fps-vs-acc} (left side). Qualitative results are shown in~\Cref{fig:qualitative} (top row). As demonstrated, accuracy-wise \methodname~is on par with other state-of-the-art methods. However, it is also clear that the results in this dataset are saturated (high-values) already, probably because its scenes are not complex and the metric is permissive~\cite{polylanenet}. This is evidenced by the small difference in performance across methods, in contrast to results in more complex datasets and less permissive metrics (as shown in~\Cref{sec:culane}). Nonetheless, our method is much faster than others.
The method proposed in~\cite{ufsa} is the only with speed comparable to ours. Since the FDR and FNR metrics were not reported in their work, we computed them using the published code and reported those metrics. Although they achieved high accuracy, the FDR is notably high. For instance, our highest false-positive rate is 5.64\%, using the ResNet-122, whereas their lowest is 18.91\%, almost four times higher.

\subsection{CULane}
\label{sec:culane}
\paragraph{Dataset description.}
CULane~\cite{scnn} is one of the largest publicly available lane detection datasets, and also one of the most complexes. All the images have $1640 \times 590$ pixels, and all test images are divided into nine categories, such as crowded, night, absence of visible lines, etc.

\paragraph{Evaluation metrics.}
The only metric is the F1, which is based on the intersection over union (IoU). Since the IoU relies on areas instead of points, a lane is represented as a thick line connecting the respective lane's points. In particular, the dataset’s official metric considers the lanes as 30-pixels-thick lines. If a prediction has an IoU greater than 0.5 with a ground-truth lane, it is considered a true positive.

\paragraph{Results.}

The results of \methodname~on CULane, along with other state-of-the-art methods, are shown in~\Cref{tab:culane} and in~\Cref{fig:fps-vs-acc} (right side). Qualitative results are shown in~\Cref{fig:qualitative} (middle row). We do not compare to the results shown in~\cite{rgconstraints}, as the main contribution is a post-processing method that could easily be incorporated to our method, but the source-code is not public. Moreover, it is remarkably slow, which makes the model impractical in real world applications (the full pipeline runs at less than 10 FPS, as reported in their work).
In this context, \methodname~achieves the highest F1 among the methods compared while maintaining a high efficiency (+170 FPS) on CULane, a dataset with highly complex scenes. Compared to~\cite{ufsa}, our most lightweight model (with ResNet-18) surpasses their largest (with ResNet-34) by almost 3\% of F1 while being much faster (250 vs. 175 FPS on the same machine). Additionally, in ``Night'' and ``Shadow'' scenes, our method outperforms all others, including SIM-CycleGAN~\cite{sim-cyclegan}, specifically designed for those scenarios. Those results demonstrate both the efficacy and the efficiency of~\methodname.

\subsection{LLAMAS}
\label{sec:llamas}
\paragraph{Dataset description.}
LLAMAS~\cite{llamas} is a large lane detection dataset with over 100k images. The annotations were not manually made, instead, they were generated using high definition maps. All images are from highway scenarios. 
The evaluation is based on the CULane's F1, which was computed by the author of the LLAMAS benchmark since the testing set's annotations are not public.%

\paragraph{Results.}
The results of \methodname~on LLAMAS, along with PolyLaneNet's~\cite{polylanenet} results, are shown in~\Cref{tab:llamas}. Qualitative results are shown in~\Cref{fig:qualitative} (bottom row).
Since the benchmark is recent and only PolyLaneNet provided the necessary source code to be evaluated on LLAMAS, it is the only comparable method. As evidenced, LaneATT is able to achieve an F1 greater than 90\% in all three backbones. The results can also be seen in the benchmark's website~\footnote{\url{https://unsupervised-llamas.com/llamas/benchmark_splines}}.

\input{tables/llamas}

\subsection{Efficiency trade-offs}
Being efficient is crucial for a lane detection model. In some cases, it might even be necessary to trade some accuracy to achieve the application's requirement. In this experiment, some of the possible trade-offs are shown. In particular, different settings of image input size ($H_I \times W_I$) and number of anchors ($N_{anc}$, as described in \Cref{met:anchor-filtering}).
The results are shown in~\Cref{tab:efficiency-tradeoffs}. They show that the number of anchors can be reduced for a slight improvement in terms of efficiency without a large F1 drop. However, if the reduction is too large, the F1 starts to drop considerably. Moreover, if too many anchors are used, the efficacy can also degrade, which might be a consequence of unnecessary anchors disturbing the training. The results are similar for the input size, although the MACs drops are larger. The largest impact of both the number of anchors and the input size is on the training time. During the inference, the proposals are filtered (using a confidence threshold) before the NMS procedure. During the training, there's no such filtering. Since the NMS is one the main bottlenecks of the model, and its running time depends directly on the number of objects, the number of anchors has a much higher impact on the training phase than on the testing phase. %

\input{tables/efficiency-tradeoffs}

\subsection{Ablation study}

This experiment evaluates the impact of each major part (one at a time) of the proposed method: anchor-based pooling, shared layers, focal loss and the attention mechanism.
The results are shown in~\Cref{tab:ablation-study}. The first row comprises the results for the standard~\methodname, while the following rows show the results for slightly modified versions of the standard model.
In the second row, the anchor-based pooling was removed and the same procedure to select features of Line-CNN~\cite{linecnn} was used (\ie, only features from a single point in the feature map were used for each anchor). In the third one, instead of using a single pair of fully-connected layers ($L_{reg}$ and $L_{cls}$) for the final prediction, three pairs (six layers) were used, one pair matching one boundary (left, bottom, or right). That is, all anchors starting in the left boundary of the image had its proposals generated by the same pair of layers $L_{reg}^L$ and $L_{cls}^L$ and similarly for the bottom ($L_{reg}^B$ and $L_{cls}^B$) and the right ($L_{reg}^R$ and $L_{cls}^R$) boundaries. In the fourth one, the Focal Loss was replaced with the Cross Entropy, and in the last one, the attention mechanism was removed.

The massive drop of performance when the anchor-based pooling procedure is removed shows its importance. This procedure enabled the use of a more lightweight backbone, which was not possible in Line-CNN~\cite{linecnn} without a large performance drop. The results also show that a layer for each boundary of the image is not only unnecessary, but also detrimental to the model's efficiency. Furthermore, using the Focal Loss instead of the Cross Entropy was also shown to be beneficial. Besides, it also eliminates the need for one hyperparameter (the number of negative samples to be used in the loss computation).
Finally, the proposed attention mechanism is another modification that significantly increases the model performance.

\input{tables/ablation-study}

%% file: tables/datasets.tex
\begin{table}[ht]
    \begin{center}
        \resizebox{\columnwidth}{!}{%
            \begin{tabular}{@{}lrrrr@{}}
            \toprule
            \textbf{Dataset}         & \textbf{Train} & \textbf{Val.} & \textbf{Test} & \textbf{Max. \# of lanes} \\ \midrule
            TuSimple~\cite{tusimple} & 3,268          & 358           & 2,782         & 5 \\
            CULane~\cite{scnn}       & 88,880         & 9,675         & 34,680        & 4 \\
            LLAMAS~\cite{llamas}     & 58,269         & 20,844        & 20,929        & 4 \\ \bottomrule
            \end{tabular}
        }
    \end{center}
    \caption{Overview of the datasets used in this work.}
    \label{tab:datasets}
\end{table}

%% file: tables/tusimple.tex
\begin{table*}
    \begin{center}
            \begin{tabular}{@{}lrrrrrr@{}}
                \toprule
                \textbf{Method}           & \textbf{F1 (\%)} & \textbf{Acc (\%)} & \textbf{FDR (\%)} & \textbf{FNR (\%)} & \textbf{FPS} & \textbf{MACs (G)} \\ \midrule
                \textbf{Source-code unavailable} &&&&&& \\ \midrule
                EL-GAN~\cite{elgan}                           & 96.26  & 94.90  & 4.12  & 3.36  & 10.0  & \\
                Line-CNN~\cite{linecnn}                       & 96.79  & 96.87  & 4.42  & 1.97  & 30.0  & \\
                FastDraw (ResNet-18)~\cite{fastdraw}          & 94.59  & 94.90  & 6.10  & 4.70  &       & \\
                PointLaneNet~\cite{pointlanenet}              & 95.07  & 96.34  & 4.67  & 5.18  & 71.0  & \\
                \cite{end2end}                                & 95.80  &        &       &       & 71.5  & \\
                R-18-E2E~\cite{e2e-lmd}                       & 96.40  & 96.04  & 3.11  & 4.09  &       & \\
                R-34-E2E~\cite{e2e-lmd}                       & 96.58  & 96.22  & 3.08  & 3.76  &       & \\
                \midrule

                \textbf{Source-code available} &&&&&& \\ \midrule
                SCNN~\cite{scnn}                              & 95.97                                & \underline{96.53}                                 & 6.17                                 & \textbf{1.80}                        & 7.5                              & \\
                Cascaded-CNN~\cite{cascaded}                  & 90.82                                & 95.24                                 & 11.97                                & 6.20                                 & 60.0                             & \\

                ENet-SAD~\cite{enet-sad}                      & 95.92                                & \textbf{96.64}                                 & 6.02                                 & \underline{2.05}                                 & 75.0                             & \\
                \cite{ufsa} (ResNet-18)                       & 87.87                                & 95.82                                 & 19.05                                & 3.92                                 & \textbf{312.5}                   & \\

                \cite{ufsa} (ResNet-34)                       & 88.02                                & 95.86                                 & 18.91                                     & 3.75                                     & 169.5                            & \\

                PolyLaneNet~\cite{polylanenet}                & 90.62                                & 93.36                                 & 9.42                                 & 9.33                                 & 115.0                       & \textbf{1.7} \\

                \midrule

                \textbf{\methodname~(ResNet-18)}              & \underline{96.71}                                & 95.57                                 & \underline{3.56}                                 & 3.01                                 & \underline{250}                              & \underline{9.3} \\

                \textbf{\methodname~(ResNet-34)}              & \textbf{96.77}                    & 95.63                                 & \textbf{3.53}                                 & 2.92                                 & 171                              & 18.0 \\

                \textbf{\methodname~(ResNet-122)}             & 96.06                                & 96.10                                 & 5.64                                 & 2.17                                 & 26                               & 70.5 \\ \bottomrule

            \end{tabular}

    \end{center}
    \caption{State-of-the-art results on TuSimple. For a fairer comparison, the FPS of the fastest method (\cite{ufsa}) was measured on the same machine and conditions as our method. Additionally, all metrics for this method were computed using the official source code, since only the accuracy was available in the paper. The best and second-best results across methods with source-code available are in bold and underlined, respectively.\looseness=-1}

    \label{tab:tusimple}

\end{table*}

%% file: tables/culane.tex
\begin{table*}

    \begin{center}

        \resizebox{\textwidth}{!}{%

            \begin{tabular}{@{}lrrrrrrrrrrrr@{}}

                \toprule

                \multicolumn{1}{c}{\textbf{Method}} & \multicolumn{1}{c}{\textbf{Total}} & \multicolumn{1}{c}{\textbf{Normal}} & \multicolumn{1}{c}{\textbf{Crowded}} & \multicolumn{1}{c}{\textbf{Dazzle}} & \multicolumn{1}{c}{\textbf{Shadow}} & \multicolumn{1}{c}{\textbf{No line}} & \multicolumn{1}{c}{\textbf{Arrow}} & \multicolumn{1}{c}{\textbf{Curve}} & \multicolumn{1}{c}{\textbf{Cross}} & \multicolumn{1}{c}{\textbf{Night}} & \multicolumn{1}{c}{\textbf{FPS}} & \multicolumn{1}{c}{\textbf{MACs (G)}} \\ \midrule
                \textbf{Source-code unavailable} &&&&&&&&&&&& \\ \midrule

                \cite{zhang2018geometric}          & 73.10                             & 89.70                              & 76.50                               & 67.40                                    & 65.50                              & 35.10                               & 82.20                             & 63.20                             &                                    & 68.70                             & 24.0                               &                                   \\

                FastDraw (ResNet-50)~\cite{fastdraw}            &                                    & 85.90                              & 63.60                               & 57.00                                    & 59.90                              & 40.60                               & 79.40                             & 65.20                             & 7013                               & 57.80                             & 90.3                           &                                   \\

                PointLaneNet~\cite{pointlanenet}    &                                    & 90.10                              &                                      &                                           &                                     &                                      &                                    &                                    &                                    &                                    & 71.0                               &                                   \\

                SpinNet~\cite{spinnet}              & 74.20                             & 90.50                              & 71.70                               & 62.00                                    & 72.90                              & 43.20                               & 85.00                             & 50.70                             &                                    & 68.10                             &                                  &                                   \\

                R-18-E2E~\cite{e2e-lmd}              & 70.80                             & 90.00                              & 69.70                               & 60.20                                    & 62.50                              & 43.20                               & 83.20                             & 70.30                             & 2296                               & 63.30                             &                                  &                                   \\

                R-34-E2E~\cite{e2e-lmd}              & 71.50                             & 90.40                              & 69.90                               & 61.50                                    & 68.10                              & 45.00                               & 83.70                             & 69.80                             & 2077                               & 63.20                             &                                  &                                   \\

                R-101-E2E~\cite{e2e-lmd}              & 71.90                             & 90.10                              & 71.20                               & 60.90                                    & 68.10                              & 44.90                               & 84.30                             & 70.20                             & 2333                               & 65.20                             &                                  &                                   \\

                ERFNet-E2E~\cite{e2e-lmd}              & 74.00                             & 91.00                              & 73.10                               & 64.50                                    & 74.10                              & 46.60                               & 85.80                             & 71.90                             & 2022                               & 67.90                             &                                  &                                   \\ \midrule

                \textbf{Source-code available} &&&&&&&&&&&& \\ \midrule

                SCNN~\cite{scnn}                    & 71.60                             & 90.60                              & 69.70                               & 58.50                                    & 66.90                              & 43.40                               & 84.10                             & 64.40                             & 1990                               & 66.10                             & 7.5                              &                                   \\

                ENet-SAD~\cite{enet-sad}                  & 70.80                             & 90.10                              & 68.80                               & 60.20                                    & 65.90                              & 41.60                               & 84.00                             & 65.70                             & 1998                               & 66.00                             & 75                               &                                   \\

                \cite{ufsa} (ResNet-18)                    & 68.40                             & 87.70                              & 66.00                               & 58.40                                    & 62.80                              & 40.20                               & 81.00                             & 57.90                             & 1743                               & 62.10                             & \textbf{322.5}                              &                                   \\

                \cite{ufsa} (ResNet-34)                    & 72.30                             & 90.70                              & 70.20                               & 59.50                                    & 69.30                              & 44.40                               & 85.70                             & \textbf{69.50}                             & 2037                               & 66.70                             & 175.0                              &                                   \\

                ERFNet-IntRA-KD~\cite{intrakd}      & 72.40                             &                                     &                                      &                                           &                                     &                                      &                                    &                                    &                                    &                                    & 100.0                              &                                   \\

                SIM-CycleGAN~\cite{sim-cyclegan}    & 73.90                             & \underline{91.80}                              & 71.80                               & 66.40                                    & 76.20                              & 46.10                               & \underline{87.80}                             & 67.10                             & 2346                               & 69.40                             &                                  &                                   \\

                CurveLanes-NAS-S~\cite{curvelane-nas}    & 71.40                             & 88.30                              & 68.60                               & 63.20                                    & 68.00                              & 47.90                               & 82.50                             & 66.00                             & 2817                               & 66.20                             &                                  & \textbf{9.0}                                  \\

                CurveLanes-NAS-M~\cite{curvelane-nas}    & 73.50                             & 90.20                              & 70.50                               & 65.90                                    & 69.30                              & 48.80                               & 85.70                             & 67.50                             & 2359                               & 68.20                             &                                  & 33.7                                  \\

                CurveLanes-NAS-L~\cite{curvelane-nas}    & 74.80                             & 90.70                              & 72.30                               & \underline{67.70}                                    & 70.10                              & \underline{49.40}                               & 85.80                             & \underline{68.40}                             & 1746                               & 68.90                             &                                  & 86.5                                  \\

                \midrule

                \textbf{\methodname~(ResNet-18)}    & 75.09                             & 91.11                              & 72.96                               & 65.72                                    & 70.91                              & 48.35                               & 85.49                             & 63.37                             & \textbf{1170}                               & 68.95                             & \underline{250}                              & \underline{9.3}                                  \\

                \textbf{\methodname~(ResNet-34)}    & \underline{76.68}                             & \textbf{92.14}                              & \underline{75.03}                               & 66.47                                    & \textbf{78.15}                              & 49.39                               & \textbf{88.38}                             & 67.72                             & 1330                               & \underline{70.72}                             & 171                              & 18.0                                  \\

                \textbf{\methodname~(ResNet-122)}   & \textbf{77.02}                             & 91.74                              & \textbf{76.16}                               & \textbf{69.47}                                    & \underline{76.31}                              & \textbf{50.46}                               & 86.29                             & 64.05                             & \underline{1264}                               & \textbf{70.81}                             & 26                               &  70.5                                 \\ \bottomrule

            \end{tabular}

        } %

    \end{center}
    \caption{State-of-the-art results on CULane. Since the images in the ``Cross'' category have no lanes, the reported number is the amount of false-positives. For a fairer comparison, we measured the FPS of the fastest method (\cite{ufsa}) under the same machine and conditions as ours. The best and second-best results across methods with source-code available are in bold and underlined, respectively.}

    \label{tab:culane}

\end{table*}

%% file: tables/llamas.tex
\begin{table}[h]
    \begin{center}
        \resizebox{\columnwidth}{!}{%
            \begin{tabular}{@{}lrrr@{}}
                \toprule
                \multicolumn{1}{c}{\textbf{Method}}           & \multicolumn{1}{c}{\textbf{F1 (\%)}} & \multicolumn{1}{c}{\textbf{Prec. (\%)}} & \multicolumn{1}{c}{\textbf{Rec. (\%)}} \\ \midrule
                PolyLaneNet~\cite{polylanenet}                & 88.40                                & 88.87                                   & 87.93 \\
                \midrule
                \textbf{\methodname~(ResNet-18)}              & 93.46                                & 96.92                                   & 90.24 \\
                \textbf{\methodname~(ResNet-34)}              & 93.74                                & 96.79                                   & 90.88 \\
                \textbf{\methodname~(ResNet-122)}             & 93.54                                & 96.82                                   & 90.47 \\\bottomrule
            \end{tabular}
        } %
    \end{center}
    \caption{State-of-the-art results on LLAMAS.}
    \label{tab:llamas}
\end{table}

%% file: tables/efficiency-tradeoffs.tex
\begin{table}
\begin{center}
\resizebox{\columnwidth}{!}{%
\begin{tabular}{@{}lrrrr@{}}
\toprule
\multicolumn{1}{c}{\textbf{Modification}} & \multicolumn{1}{c}{\textbf{F1 (\%)}} & \multicolumn{1}{c}{\textbf{FPS}} & \multicolumn{1}{c}{\textbf{MACs (G)}} & \multicolumn{1}{c}{\textbf{TT (h)}} \\ \midrule
$N_{anc}=250$                  & 68.68                           & 196                              & 17.3                              & 5.7 \\
$N_{anc}=500$                  & 75.45                           & 190                              & 17.4                              & 6.4 \\
$N_{anc}=750$                  & 75.80                           & 181                              & 17.7                              & 7.8 \\
$N_{anc}=1000$                 & 76.66                           & 171                              & 18.0                              & 11.1 \\
$N_{anc}=1250$                 & 75.91                           & 156                              & 18.4                              & 11.5 \\ \midrule
$H_I \times W_I = 180 \times 320$         & 66.74                           & 195                              & 4.8                               & 4.3\\
$H_I \times W_I = 288 \times 512$         & 75.02                           & 186                              & 11.5                              & 7.3 \\
$H_I \times W_I = 360 \times 640$         & 76.66                           & 171                              & 18.0                              & 11.1 \\ \bottomrule
\end{tabular}
}
\end{center}
\caption{Efficiency trade-offs on CULane using the ResNet-34 backbone. ``TT'' stands for training time in hours.}
\label{tab:efficiency-tradeoffs}
\end{table}

%% file: tables/ablation-study.tex
\begin{table}
\begin{center}
\resizebox{\columnwidth}{!}{%
\begin{tabular}{@{}lrrr@{}}
\toprule
\multicolumn{1}{c}{\textbf{Model}} & \multicolumn{1}{c}{\textbf{F1 (\%)}} & \multicolumn{1}{c}{\textbf{FPS}}  & \multicolumn{1}{c}{\textbf{Params. (M)}} \\ \midrule
\methodname~(ResNet-34)            & 76.68                                & 171                               & 22.13 \\
$-$ anchor-based pooling           & 64.89                                & 188                               & 21.39 \\
$-$ shared layers                  & 75.45                                & 142                               & 22.34 \\
$-$ focal loss                     & 75.54                                & 171                               & 22.13 \\
$-$ attention mechanism            & 75.78                                & 196                               & 21.37 \\ \bottomrule
\end{tabular}
}
\end{center}
\caption{Ablation study results on CULane.}
\label{tab:ablation-study}
\end{table}

%% file: secs/5_conclusion.tex
\section{Conclusion}
We proposed a real-time single-stage deep lane detection model that outperforms state-of-the-art models, as shown by an extensive comparison with the literature. The model is not only effective but also efficient. On TuSimple, the method achieves the second-highest reported F1 (a difference of only 0.02\%) while being much faster than the top-F1 method (171 vs. 30 FPS). On CULane, one of the largest and most complex lane detection datasets, the method establishes a new state-of-the-art among real-time methods in terms of both speed and accuracy (+4.38\% of F1 compared to the state-of-the-art method with a similar speed of around 170 FPS). Additionally, the method achieved a high F1 (+93\%) on the LLAMAS benchmark on all three backbones evaluated. To achieve those results, along with other modifications, a novel anchor-based attention mechanism was also proposed. The ablation study showed that this addition increased the model's performance (F1 score) significantly when considering the gains obtained by the literature advance in recent years. Additionally, some efficiency trade-offs that are useful in practice were also shown. %